\documentclass[10pt,twocolumn,letterpaper]{article}

\usepackage{cvpr}
\usepackage{times}
\usepackage{epsfig}
\usepackage{graphicx}
\usepackage{amsmath}
\usepackage{amssymb}
\usepackage{url}
\usepackage{enumitem}

\usepackage[ruled]{algorithm2e}
\usepackage{wrapfig}
\usepackage{float}
\usepackage{sidecap}
\usepackage{color}

\usepackage[pagebackref=true,breaklinks=true,letterpaper=true,colorlinks,bookmarks=false]{hyperref}

\cvprfinalcopy 

\def\cvprPaperID{1897} 

\ifcvprfinal\fi
\begin{document}

\title{Object Detection via Aspect Ratio and Context Aware Region-based Convolutional Networks}

\author{Bo Li$^{1}$, Tianfu Wu$^{3,}$\thanks{Tianfu Wu is the corresponding author.}\ , Shao Shuai$^{1,2,}$\thanks{This work was done when Shao Shuai was an Intern at YunOS BU, Alibaba Group.} \ , Lun Zhang$^{1}$ and Rufeng Chu$^{1}$ \\
$^{1}$YunOS BU, Alibaba Group \ \ \ \ \ \ \ $^{2}$Jilin University\\
$^{3}$Department of Electrical and Computer Engineering and the Visual Narrative Cluster, \\North Carolina State University  \\
\small{\it \{shize.lb,lun.zhangl, rufeng.churf\}@alibaba-inc.com, tianfu\_wu@ncsu.edu, shaoshuai2113@mails.jlu.edu.cn}
}

\maketitle

\begin{abstract}
\vspace{-5mm}
Jointly integrating aspect ratio and context has been extensively studied and shown performance improvement in traditional object detection systems such as the DPMs. It, however, has been largely ignored in deep neural network based detection systems. This paper presents a method of integrating a mixture of object models and region-based convolutional networks for accurate object detection.  Each mixture component accounts for both object aspect ratio and multi-scale contextual information explicitly:
(i) it exploits a mixture of tiling configurations in the RoI pooling to remedy the warping artifacts caused by a single type RoI pooling (e.g., with equally-sized 7 $\times$ 7 cells), and to respect the underlying object shapes more; (ii) it ``looks from both the inside and the outside of a RoI" by incorporating contextual information at two scales: global context pooled from the whole image  and local context pooled from the surrounding of a RoI. To facilitate accurate detection, this paper proposes a multi-stage detection scheme for integrating the mixture of object models, which utilizes the detection results of the model at the previous stage as the proposals for the current in both training and testing.
The proposed method is called the aspect ratio and context aware region-based convolutional network (ARC-R-CNN).   
In experiments, ARC-R-CNN shows very competitive results with Faster R-CNN~\cite{faster_rcnn} and R-FCN~\cite{rfcn} on two datasets: the PASCAL VOC and the Microsoft COCO. It obtains significantly better mAP performance using high IoU thresholds on both datasets.  
\end{abstract}

\vspace{-5mm}
\section{Introduction}

\vspace{-1mm}
\subsection{Motivation and Objective}
\vspace{-1mm}
\begin{figure}
\centering
{\includegraphics[width = 0.45\textwidth]{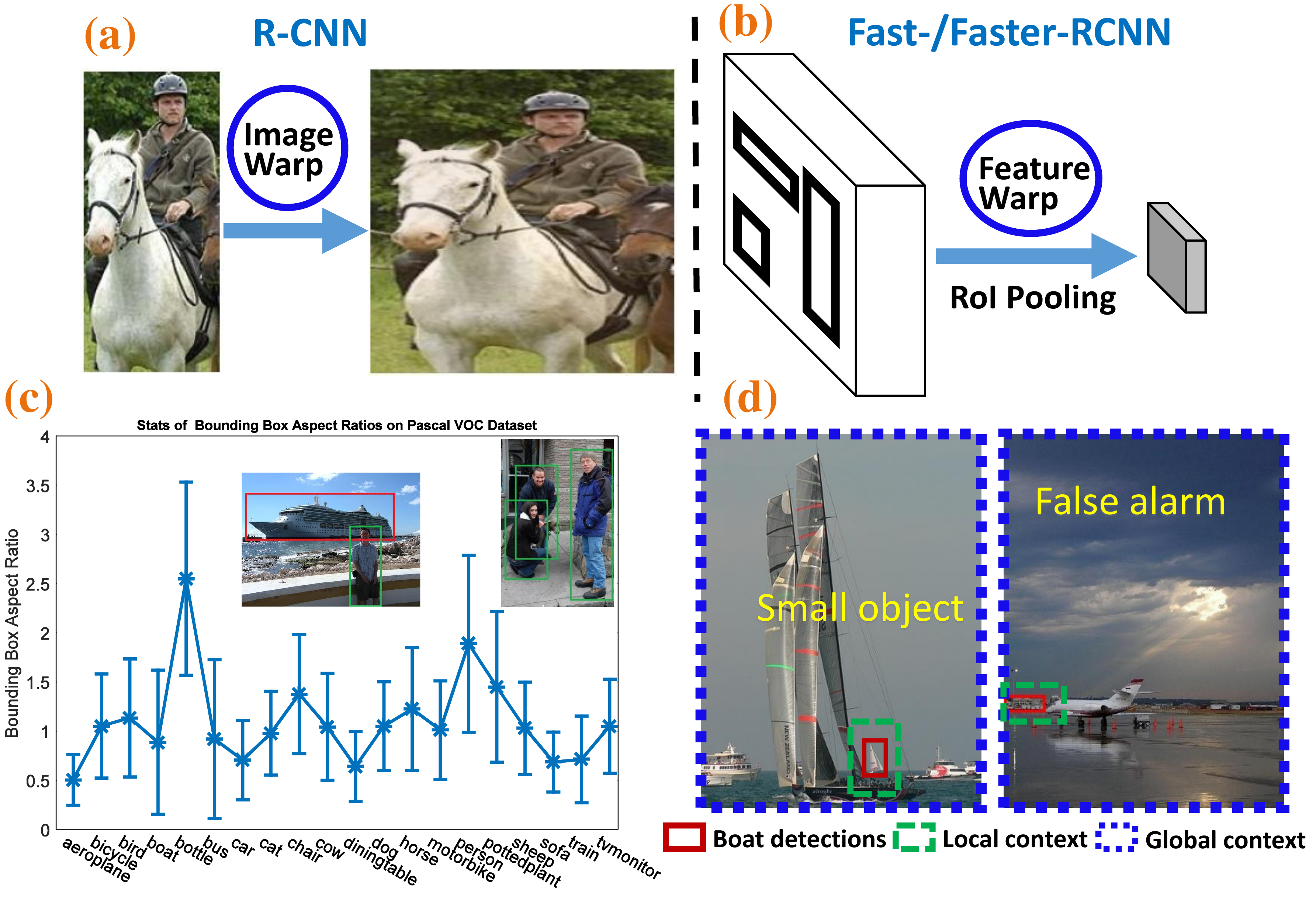}}
\caption{Motivation of the proposed method. It remedies the warping artifacts in state-of-the-art R-CNN object detection systems (see (a) and (b)) by integrating aspect ratio in the RoI pooling to respect the underlying distribution of aspect ratios  (see plots and examples in (c)). It also takes into account both global and local contextual information and thus is more accurate by improving small object detection and by suppressing false alarms (see the example of boat detection in (d)). See text for details. \vspace{-5mm}
}
\label{fig:demo}
\end{figure}

We have witnessed a critical shift in the literature of object detection from explicit models such as the mixture of deformable part-based models (DPMs)~\cite{DPM} and its many variants, and hierarchical and compositional And-Or graph models~\cite{carAOG}, to less transparent but much more accurate deep neural network based approaches~\cite{lecun89,rcnn,fast_rcnn,faster_rcnn,resNet,googlenet,yolo,ssd,rfcn}. A popular framework of deep neural networks in object detection is the region-based convolutional neural networks (R-CNN)~\cite{rcnn}, which consists of two components: (i) A region proposal component is used to reduce the number of candidates to be classified. 
The proposals are generated  either by utilizing off-the-shelf objectness detectors such as the selective search~\cite{SS}, Edge Boxes \cite{edge_boxes} and BING~\cite{BING} or by learning an integrated region proposal network (RPN)~\cite{faster_rcnn} in an end-to-end way.
(ii) A prediction component is used to classify all the proposals and regress  bounding boxes of the classified object candidates. To accommodate different shapes and sizes of the proposals, either raw image patches are warped to the same canonical size as done in the R-CNN~\cite{rcnn} (see Fig.~\ref{fig:demo} (a)) or later on a more effective feature warping operator is designed based on the region-of-interest (RoI) pooling~\cite{fast_rcnn} (see Fig.~\ref{fig:demo} (b)) to compute equally-sized feature maps. These warping artifacts are purely caused by design choices for the simplicity of applying deep neural networks in detection and for the practical consideration of affordable training and testing time complexities. They are less elegant than traditional explicit models in term of respecting the underlying distribution of object shapes (see Fig.~\ref{fig:demo} (c)), although they obtain much better performance. In the literature, Girshick et al.~\cite{dpdpm} and Wan et al.~\cite{wanli} have made some progress on integrating DPMs and convolutional networks, but obtained significantly lower average precision than R-CNN.

\textit{This paper is motivated by two straightforward and intuitive questions: What are there that have shown performance improvements in the traditional object detection systems, but are largely ignored in the R-CNN based detection framework? And, would they also improve the performance of state-of-the-art R-CNN object detection systems if integrated properly?} 
Among many others, this paper addresses two issues jointly: one is to remedy the feature warping artifacts by accounting for model aspect ratios in RoI pooling explicitly. Considering the aspect ratios of diverse objects helps localizing them more accurately. 
The other is to take into account multi-scale contextual information while most of the deep neural network based detection systems take advantage of global context (from the whole image) only. Fig.~\ref{fig:demo} (d) illustrates the different roles played by local and global context. We can see the whole image of a airport scenario is quite similar to that of a sea scene. In this case, global context might be confused, but the local context will be still helpful as it shows surrounding the object is building in an airport and thus suppresses a boat detection.

To facilitate accurate object detection, this paper uses a multi-stage detection scheme in the spirit similar to the multi-stage method for human pose recovery in~\cite{cao2016realtime}, which consists of multiple stages of the proposed aspect ratio and context aware mixture of object models.  

\begin{figure} 
\centering
{\includegraphics[width = 0.5\textwidth]{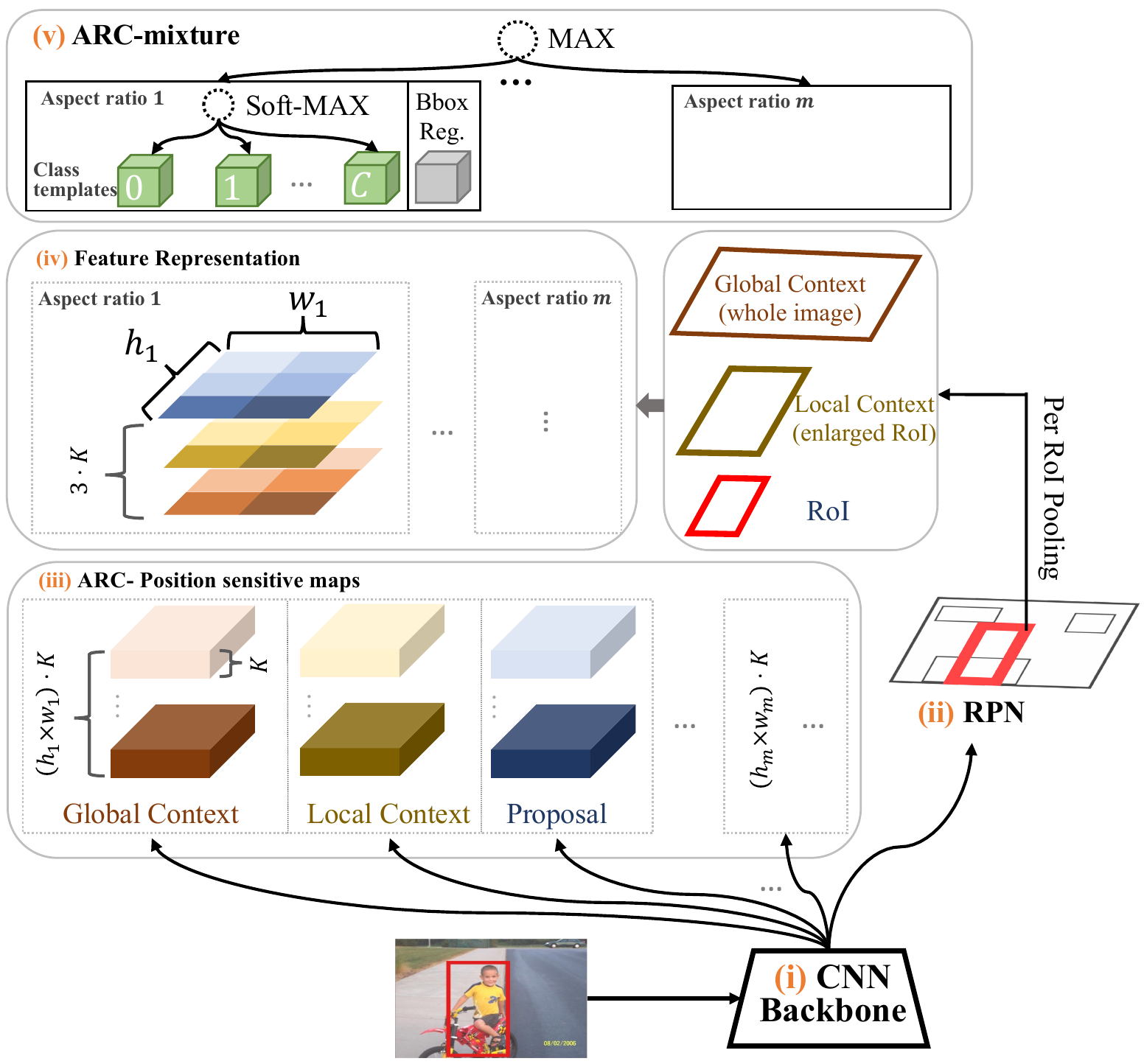}}
\caption{Illustration of the proposed method. It integrates a mixture of aspect ratio and context aware object models (including class score prediction and bounding box regression) with state-of-the-art region-based convolutional networks~\cite{faster_rcnn,rfcn}. See text for details. (Best viewed in magnification and color.) 
}
\vspace{-3mm} 
\label{fig:model}
\end{figure} 

\vspace{-1mm}
\subsection{Method Overview} 
\vspace{-1mm}
This paper presents an end-to-end integration between a mixture of aspect ratio and context aware object models and state-of-the-art R-CNNs, which is dubbed the ARC-R-CNN.
Fig.~\ref{fig:model} shows the overall architecture. 
In the spirit similar to Faster R-CNN~\cite{faster_rcnn} and R-FCN~\cite{rfcn}, object detection via ARC-R-CNN in an image consists of two stages: generating category-agnostic bounding box proposals (or so-called objectness detection) and classifying each proposal into one of the $C+1$ categories (e.g., the $C=20$ classes in the PASCAL VOC dataset and a background class $0$) in terms of prediction scores with the bounding box regressed in a category-agnostic way for simplicity. The former is implemented by the RPN component and the latter is done through the RoI-wise prediction based on the ARC-mixture model.
The proposed ARC-R-CNN consists of five components: 
\begin{itemize} [leftmargin=*]
\itemsep0em 
\item [(i)] A fully convolutional network backbone using state-of-the-art network architecture such as the residual net~\cite{resNet} (with the average pooling layer and fully connected layer removed), which is shared by the two stages above.
\item [(ii)] A region proposal network (RPN) which computes proposal RoIs in a category-agnostic way (so called objectness detection~\cite{SS}), as done in faster R-CNN~\cite{faster_rcnn}.
\item [(iii)] ARC-position-sensitive maps which are built on top of the position-sensitive map proposed in the R-FCN~\cite{rfcn}. For each aspect ratio (i.e., a tiling configuration to be used in RoI pooling, e.g., the first aspect ratio defined by $h_1\times w_1$), we have three maps specific to global context, local context and bounding box proposals respectively; The three maps are said to be position-sensitive in the sense that their channels (e.g., $(h_1\times w_1)\times K$) are divided into $h_1\times w_1$ subsets, and the first subset ($K$ channels) is used by the first cell indexed by $(0,0)$ out of the $h_1\times w_1$ cells in RoI pooling based on the first aspect ratio in pooling, and the second subset by the second cell indexed by $(0,1)$, and so on so forth.   
\item [(iv)] A per RoI pooling layer which extracts features from the ARC-position-sensitive maps.  For each proposal RoI computed by the RPN, the corresponding local context RoI is centered at the center of the proposal RoI with enlarged size, and the global context RoI uses the whole map. Then, the three RoIs are interpreted in terms of $m$ aspect ratios. For each aspect ratio, the features are extracted from the proposal, local context and global context specific position-sensitive maps, and then concatenated together (e.g., to form a $3\cdot K\times h_1 \times w_1$ feature map for the first aspect ratio).  
\item [(v)] A mixture of $m$ ARC-aware object models, each of which computes $(C+1)$-class prediction scores using $(C+1)$-class object templates w.r.t. soft-max, and a category-agnostic $4$-$d$ bounding box regression vector. For example, consider the first aspect ratio component, each of the object template is parameterized by a tensor of shape $3\cdot K\times h_1 \times w_1$. The final output is computed w.r.t. a MAX operator which selects the best aspect ratio component based on the soft-max scores of the predicted class from the $m$ components. \textbf{The multi-stage detection scheme} utilizes a cascade of the mixture: the first stage uses the RPN RoIs as proposals, and the $i$-th stage uses the detection results of the $i-1$-th stage as proposals ($i>1$). The cascade improves the accuracy at high IoU thresholds significantly (2-stage is used in experiments and see ablation studies in Section~\ref{sec:training}).    
\end{itemize}

The ARC-R-CNN is trained end-to-end under the multi-task (classification and bounding box regression) setting same as the Fast/Faster R-CNN~\cite{fast_rcnn,faster_rcnn}. 


We note that we use class-agnostic aspect ratios and simple contextual patterns for simplicity. Strictly speaking we do not eliminate the feature warping artifacts. The proposed method is, however, generic and can be easily extended to handle class-specific aspect ratios at the cost of model complexity and training and testing time.

In experiments, we test our model on the Pascal VOC and MS COCO datasets~\cite{pascal}. Using the 50-layer and 101-layer Residual Net (ResNet-50 and ResNet-101)~\cite{resNet} as the backbone, our ARC-R-CNN yields consistently better performance in terms of mAP (especially for high IoU thresholds) than the Faster R-CNN~\cite{faster_rcnn} and R-FCN~\cite{rfcn}. 

\textit{Remarks:} By treating tiling cells in a RoI as parts, the proposed method can further be interpreted as an end-to-end integration between R-CNN~\cite{fast_rcnn,faster_rcnn} and a mixture of ARC-aware part-based models, and hopefully sheds light on building explainable R-CNN based object detection systems ~\cite{XAI} by integrating more explicit hierarchical models such as the grammar models~\cite{zhu_grammar,pff_grammar,carAOG,leozhu_nips07,DisAOT_CVPR2013}.

\vspace{-2mm}
\section{Related Work and Our Contributions}
 \vspace{-1mm}

\textbf{CNN-based Detection Models.}
Object detection has been greatly improved by convolutional neural networks (CNN) on both accuracy and speed \cite{szegedy_cvpr14,szegedy_nips13,rcnn, overfeat,rcnn,fast_rcnn,faster_rcnn,rfcn,yolo,ssd}. Currently, there are two most influential research streams: a) along the first stream is region based detection models \cite{rcnn,sppnet,fast_rcnn,faster_rcnn,rfcn}, which consisting of a region proposal module and a recognition module. Those works all rely on pre-computed object proposals to predict the position of each underlying object; b) along the second stream is the sliding window based models \cite{yolo,ssd}, which predicting object positions directly in a sliding window manner. Though sliding window based cnn detectors (e.g., YOLO, SSD) are faster than region based R-CNN style models, the R-CNN style models seems more accurate, as they still perform leading accuracies on popular benchmarks \cite{pascal,imagenet,coco}.

R-CNN style models often use either image warp \cite{rcnn,yuting,yukun}, or feature warp \cite{sppnet, fast_rcnn,faster_rcnn} to transform the whole image feature to per RoI feature. Usually, the transformed per RoI feature are equally-sized. Though simplicity, the choice of those warping artifacts is a compromise of effectiveness and affordable computational costs. As studied in \cite{wanli}, this is less elegant than traditional models, e.g., DPM \cite{DPM,pff_grammar,mixture_dpm,ssdpm}, AOG \cite{zhu_grammar,carAOG}.

\textbf{Mixture Modeling.}
Many traditional works \cite{DPM,pff_grammar,mixture_dpm,ssdpm,ramananPose,subcat,3dvp,Lopez2011,bojan_cvpr13,carAOG} explicitly model the intra-class or sub-category variations by mixture modeling. 
Specifically, \cite{DPM,pff_grammar} model mixtures by the aspect ratio of object bounding box. This is a simple yet effective way to quantify the visual space. \cite{Lopez2011} models mixtures by 2D viewpoints, \cite{bojan_cvpr12,mono3d} model mixtures by 3D viewpoints, \cite{bojan_cvpr13,xiaogang1,carAOG} model mixtures by occlusion extent of an object. In \cite{mixture_dpm,subcat,3dvp}, mixtures are discovered by feature clustering. In \cite{ramananPose,xianjie}, mixtures are modeled based on the relative distance of neighboring joints.


\textbf{Contextual Modeling.}
The role of context has been well exploited in recognition and detection \cite{desai, carAOG, guangchen_cvpr, hoiem06, torralba, haoxiang, laptev15}. For generic object detection with CNN,
\cite{mrcnn, multipath} utilized multi-scale local context to improve the localization of objects.
For ResNets-based Faster R-CNN model \cite{resNet}, only global context was utilized, and it introduced expensive computational costs. In this paper, we provide an efficient and effective way to jointly integrate local and global context.

\textbf{Accurate Localization.}
Many works resort to auxiliary features to aid object localization. \cite{fidler} incorporated segmentation cues with DPM, \cite{hoiem_loc} utilized color and edge features, \cite{Schulter_2014_CVPR} used the height prior of an object. To improve the localization ability of R-CNN,
\cite{yuting} used Bayesian optimization to refine the bounding box proposals and trained the CNNs with a structured loss. \cite{locNet} assigned probabilities on each row and column of a search region to get accurate object positions. Those works are complementary with our framework.


This paper makes the following main contributions to object detection. 

(i) It proposes a simple yet effective framework to integrate explicit models (e.g., a mixture of category-agnostic aspect ratio and context aware object models) with R-CNN.

(ii) It proposes an effective and efficient way to integrate aspect ratio, context, and position sensitive RoI pooling to bridge top-down explicit models and bottom-up powerful deep learning based features.

(iii) It provides a simple yet effective multi-stage detection scheme for accurate object detection.

(iv) It obtains state-of-the-art detection performance on the PASCAL VOC 2007 and 2012, and MS COCO datasets w.r.t. Faster R-CNN~\cite{faster_rcnn} and R-FCN~\cite{rfcn}.

\vspace{-2mm}
\section{The Proposed Model} \label{sec:model}
\vspace{-1mm}
In this section, we present details of the proposed ARC-R-CNN (see Fig.~\ref{fig:model}). 
Denote by $\Lambda$ an image lattice and by $I_{\Lambda}$ an image defined on the lattice $\Lambda$. We formulate the two stages in object detection via ARC-R-CNN in the following. 

\textbf{The RPN} subnetwork, denoted by $r(I_{\Lambda};\Theta_{RPN})$,  computes a set of category-agnostic bounding box proposals (foreground vs background),
\begin{equation}
r(I_{\Lambda};\Theta_{RPN})=\{(B_i, t_i, l_i, p_i); i=1,\cdots, M\}\label{eqn:RPNscoring}
\end{equation}  
where $\Theta_{RPN}$ is the parameters including parameters in the convolution network backbone, $l_i\in\{0, 1\}$, $p_i$ the prediction probability and $t_i=(t_i^x, t_i^y, t_i^{wd}, t_i^{ht})$ a 4-d vector of bounding box regression parameters used to refine $B_i$. The total number $M$ is determined by the size of the feature map of the last layer in the convolutional network backbone (see Fig.~\ref{fig:model}) and the number of translation-invariant anchor boxes (details are referred to the Faster R-CNN~\cite{faster_rcnn}). For a pair of $(B, t)$, let $B=(x, y, wd, ht)\subseteq \Lambda$ where $(x,y)$ represents the pixel coordinates of the bounding box center and $(wd, ht)$ the width and height respectively. The refined bounding box $B'$ is computed following the parameterization in Fast R-CNN~\cite{fast_rcnn}: $B'=(x', y', wd', ht')$ where  
$x' = t^x \cdot wd + x$, $y' = t^y \cdot ht + y$, $wd' = wd \cdot \exp(t^{wd})$ and $ht' = ht \cdot \exp(t^{ht})$.

Before feeding into the prediction subnetwork, the set of foreground proposals (i.e., $l_i=1$) is pruned by non-maximum suppression (NMS) with a predefined intersection-over-union (IoU) threshold $\tau_{RPN}$.  For notional simplicity, we still denote by $B=(x, y, wd, ht)$ a proposal RoI after regression and NMS. 

\textbf{The category-agnostic mixture} consists of a small number $m$ of aspect ratio components each of which represents a tiling configuration of equally-sized $h_i\times w_i$ cells in RoI pooling ($i=1, \cdots, m$). The specification of aspect ratios takes into account object shape statistics in training data of all categories (see the ablation studies in the experiments). For a proposal RoI $B$, each mixture component computes a $(C+1)$-$d$ class score vector, $S_i=(s^i_0,s^i_1, \cdots, s^i_{C})$ using $C+1$ object class templates (where we omit the index of $B$ in $S_i$ for notational clarity and without confusion) and a category-agnostic $4$-$d$ bounding box regression vector for simplicity. The best aspect ratio component is selected by a MAX operator on the class scores: $i^*=\arg\max_i \max_{j\in[0,C]} s^i_j$ and its regression parameters are used to further refine the proposal RoI $B$ in the same way as stated above. The final predicted class label for the proposal RoI $B$ is $\hat{y}=\arg\max_j s^{i^*}_j$.

To compute the class score vector and the bounding box regression vector for each aspect ratio component, we first build three ARC-position sensitive maps from the last layer of CNN backbone in the spirit similar to R-FCN~\cite{rfcn}. The three maps are shared among all proposal RoIs for extracting features w.r.t. global context, local context and a proposal RoI itself respectively in the RoI pooling with the tiling configuration defined by the aspect ratio. 

\textit{The ARC-position-sensitive score maps.} For an aspect ratio component $i$, denote by $\mathbb{F}^i_{roi}$, $\mathbb{F}^i_{local}$ and $\mathbb{F}^i_{global}$ the three position-sensitive maps for the proposal RoI itself,  the local context and the global context respectively. The three maps have the same dimensions: the width $W$ and the height $H$ (they are the same as those of the last layer in the CNN backbone), and the number of channels is $K\times h_i\times w_i$ since we have $h_i\times w_i$ cell positions due to the tiling based on the aspect ratio and each cell position accounts for a predefined number $K$ of channels. 

\textit{The RoI pooling w.r.t. an aspect ratio component $i$.} Given a proposal RoI $B$, denote by $b_{roi}$, $b_{local}$ and $b_{global}$ the transformed bounding boxes of $B$ in the three feature maps ($\mathbb{F}^i_{roi}$,  $\mathbb{F}^i_{local}$ and $\mathbb{F}^i_{global}$) respectively, which map the RoI from the pixel coordinates to the coordinates in the position-sensitive maps. $b_{local}$ is centered at the same position as $b_{roi}$ with the side length enlarged by a factor $\lambda$ ($\lambda=1.5$ in our experiments). $b_{global}=(0, 0, W, H)$ is the same for all proposal RoIs. Th same RoI pooling procedure is done for the three bounding boxes $b_{roi}$, $b_{local}$ and $b_{global}$. Without loss of generality and notional confusions, consider the pooling in a transformed bounding box $b=(x,y,wd,ht)$ with left-top coordinates $(x,y)$, width $wd$ and height $ht$, we first tile $b$ into $h_i\times w_i$ cells.  For each cell $b_{j,k}=(x+(k-1)\lfloor \frac{wd}{w_i} \rfloor, y+(j-1)\lfloor \frac{ht}{h_i}\rfloor, \lfloor \frac{wd}{w_i} \rfloor, \lfloor \frac{ht}{h_i}\rfloor)$ where $1\leq j \leq h_i, 1\leq k \leq w_i$ and the height of cells at the bottom row and the width of cells at the right-most column will compensate the rounding effect accordingly, a $K$-$d$ feature vector is computed using average pooling from the corresponding ARC-position-sensitive map, as done in the R-FCN~\cite{rfcn}. So, each cell is described by a $3\cdot K$-$d$ feature vector and the feature map for a proposal RoI $B$, denoted by $f^i_B$, is with shape $3\cdot K\times h_i \times w_i$.

Each mixture component consists of $C+1$ ARC-aware object templates and a category-agnostic bounding box regressor.  Denote by $\Theta^{i,c}_{ARC}$ the set of template parameters of the class $c \in [0, C]$ in  the $i$-th aspect ratio mixture component, and all the $C+1$ templates have the same shape $3\cdot K \times h_i \times w_i$ corresponding to the extracted feature map above.  We have the class score $s_c^i$ computed by, 
\begin{equation}
s_c^i = <f^i_B, \Theta^{i,c}_{ARC}>
\end{equation}
Denote by $\Theta^{i, reg}_{ARC}$ the set of parameters of category-agnostic bounding box regressor, which is represented by a tensor with shape $4\times (3\cdot K \times h_i \times w_i)$, and the $4$-$d$ regression vector is also computed by inner product between $f^i_B$ and $\Theta^{i, reg}_{ARC}$. 
We note that in the architecture stated above each mixture component can be implemented by a fully connected layer with the number of outputs being $(C+1)+4$ on top of the extracted features.

\vspace{-2mm}
\section{Parameter Learning by Iterative Training\label{sec:training}}
\vspace{-1mm}

In this section, we briefly present the multi-task formulation in learning parameters, which are the same as in Fast/Faster R-CNN~\cite{fast_rcnn,faster_rcnn}.  Then we present an effective iterative training procedure, as well as some implementation details.

\textbf{Multi-Task Training for Subnetworks.}
Both the RPN subnetwork and the ARC-R-CNN subnetwork is trained using a multi-task loss including the classification loss and the bounding box regression loss as in \cite{fast_rcnn,faster_rcnn}. The objective function is defined as follows:
\begin{align}
\nonumber \mathcal{L}(\{p_j, \ell_j\}, \{t_j, t_j^*\},) = &{1\over N_{cls}}\sum_j L_{cls}(p_j, \ell_j) + \\
& \lambda {1\over N_{reg}}\sum_j 1_{\ell_j\geq 1}L_{reg}(t_j, t_j^*)
\end{align}
where $j$ is the index of an anchor (for training the RPN with the ground-truth label $\ell_j\in\{0, 1\}$) or a proposal RoI (for training the ARC-R-CNN with $\ell_j\in [0, C]$) in a mini-batch, $p_j$ the predicted probability (i.e., soft-max score) of the anchor or the RoI being a category $\ell_j$ and $L_{cls}(p_j, \ell_j)=-\log p_j$, $t_j$ and $t_j^*$ the predicted  $4$-$d$ bounding box regression vector and and ground-truth one respectively and $L_{reg}(t_j, t_j^*)$ uses the smooth-$L_1$ loss proposed in~\cite{fast_rcnn}.  The term $1_{\ell_j\geq 1}$ means that we only take into account the bounding box regression loss of positives. $N_{cls}$ is usually set to the size of mini-batch and $N_{reg}$ the number of anchor positions in training RPN and the number of proposal RoIs in training ARC-RCNN. $\lambda$ is a trade-off parameter to balance the two types of loss.

\textbf{Iterative Training for Multi-Stage Detection Scheme.} 
We use a cascade of 2-stage ARC-R-CNN. The overall procedure is a straightforward extension of the alternating training for the Faster R-CNN~\cite{faster_rcnn}. It consists of five steps:  
 1) Train the initial RPN subnetwork with an ImageNet pre-trained backbone and generate initial propsoal RoIs; 2) Train the initial first stage ARC-R-FCN subnetwork with the same ImageNet pre-trained backbone using the initial proposal RoIs; The CNN backbone is then fixed and shared between the RPN and the  ARC-R-CNN. 3) Retrain the RPN and generate new proposal RoIs; 4) Retrain the first stage ARC-R-RCNN with the new proposal RoIs and generate the detection results with high recall kept; 5) Train the second stage ARC-R-RCNN using the first stage detection results as proposal RoIs. We can add more stages following the same procedure. We found that the 2-stage cascade works very well in experiments and thus we do not try more stages to reduce the model complexity and the training/testing time. 


\textit{Implementation details.} ARC-R-CNN is implemented using the open source Caffe library \cite{caffe}. OHEM \cite{ohem} is utilized for efficient training, $128$ RoIs are selected for backpropagation \cite{rfcn}. To cope with small objects, the $\grave{A}$ $trous$ algorithm \cite{hole, mallat} is utilized to enlarge the last convolutional feature maps in ResNets.
Training images are resized such that the min scale (shorter side of image) is $600$ pixels, and the max scale (longer side of image) is $1000$ pixels as \cite{fast_rcnn}.  
For each training step, we use a weight decay of $0.0005$ and a momentum of $0.9$. 
In this paper, we focus on investigating the role of aspect ratios and context.
For simplicity, we don't use other training and testing tricks, e.g., multi-scale training/testing, test with left-right flipped images \cite{ion}.

\vspace{-2mm}
\section{Experiments}
\vspace{-1mm}

\subsection{Pascal VOC Datasets}
\vspace{-1mm}
\textbf{PASCAL VOC 2007 Testset}. We first verify our method on the PASCAL VOC 2007 dataset \cite{pascal}. Following \cite{faster_rcnn,rfcn}, the union set of VOC 2007 \textit{trainval} and VOC 2012 \textit{trainval} (``07+12") are used for training, and the VOC 2007 \textit{test} set is used for testing. 
During testing, non-maximum supression (NMS) is used to report the final results.
We adopt the PASCAL VOC evaluation protocol \cite{pascal}, i.e., a detection is correct only if the intersection over union (IoU) of its bounding box and the groundtruth bounding box are equal or greater than $0.5$, and evaluate our model by mean average precision (mAP). 
In this experiment, ARC-R-CNN is finetuned using a learning rate of $0.001$ for the first $80k$ iterations and $0.0001$ for another $30k$ iterations with a mini-batch size of $2$. 
To cover most of aspect ratios among the PASCAL $20$ object classes, we use aspect ratios as $\{7\times7, 5\times10, 10\times5, 4\times12, 12\times4,3\times12, 12\times3\}$ for ARC-R-CNN-Res50s, and $\{7\times7, 5\times10, 10\times5, 3\times12, 12\times3\}$ for ARC-R-CNN-Res101s.

\begin{table} 
\begin{center}
\resizebox{0.95\hsize}{!}{
\begin{tabular}{|c|c|c|c|c|}
\hline
\multicolumn{4}{|c|}{PASCAL VOC 2007 (IoU $\ge 0.5$)}  \\
\hline
Method & training data & test times & mAP\\
\hline
R-FCN-Res50 \cite{rfcn} & $07$+$12$ & $0.12$ & $77.4$  \\
R-FCN-Res50-ReIm & $07$+$12$ & $0.14$ & $77.3$  \\
ARC-R-CNN-Res50 & $07$+$12$ & $0.23$ & $\mathbf{80.2}$  \\
\hline
Faster-RCNN-Res101 \cite{faster_rcnn} & $07$+$12$ & $0.42$ & $76.4$  \\
R-FCN-Res101 & $07$+$12$ & $0.17$ & $79.5$  \\
R-FCN-Res101-ReIm & $07$+$12$ & $0.20$ & $79.4$  \\
ARC-R-CNN-Res101 & $07$+$12$ & $0.38$ & $\mathbf{82.0}$  \\
\hline
\end{tabular}
}
\end{center}
\caption{mAP results with IoU $\ge 0.5$ of Faster-RCNN, R-FCN and ARC-R-CNN on PASCAL VOC 2007 test set. ResNet-50 and ResNet-101 are used as the backbone architectures. Time is evaluated on a Nvidia K40 GPU.}
\label{tab:07coarse} 
\vspace{-3mm}
\end{table}

Table \ref{tab:07coarse} shows the IoU $\ge 0.5$ results of our model and state-of-the-art Faster R-CNN \cite{faster_rcnn} and R-FCN \cite{rfcn} models. For fairness, Faster R-CNN and R-FCN are also compared without using MS COCO data and multi-scale training.
For the convenience of comparison with R-FCN on bounding box localization under the same setting, we also reimplement two R-FCN models as R-FCN-Res50-ReIm and R-FCN-Res101-ReIm. The mAP results are consistent with the original paper \cite{rfcn}.
From Table \ref{tab:07coarse}, we can see our models outperform R-FCNs by $2.8$ and $2.5$ points in terms of mAP with ResNet-50 and ResNet-101, respectively. Besides, it also surpasses Faster-RCNN with ResNet-101 by $5.6$ points. These results verify the superiority of ARC-R-FCN with integrating aspect ratio, context and iterative training.

\begin{table*} 
\begin{center}
\resizebox{0.97\hsize}{!}{
\begin{tabular}{|c|c|c|c|c|c|c|c|}
\hline
\multicolumn{8}{|c|}{PASCAL VOC 2007 (IoU $\ge 0.7$)}  \\
\hline
Method & R-FCN-Res50-ReIm & R-FCN-Res101-ReIm & R-FCN-Res101 \cite{rfcn} & \cite{yuting} & LocNet \cite{locNet} & ARC-R-CNN-Res50 & ARC-R-CNN-Res101 \\
\hline
mAP & $57.8$ & $60.5$ & $60.5$ & $43.7$ & $65.4$ & $64.7$ & $\mathbf{68.2}$ \\
\hline
\end{tabular}
}
\end{center}
\caption{mAP results with IoU $\ge 0.7$ of ARC-R-CNN and state-of-the-art models on PASCAL VOC 2007 test set.}
\label{tab:07fine} 
\vspace{-3mm}
\end{table*}

\begin{figure*}
\centering
{\includegraphics[width = 0.95\textwidth]{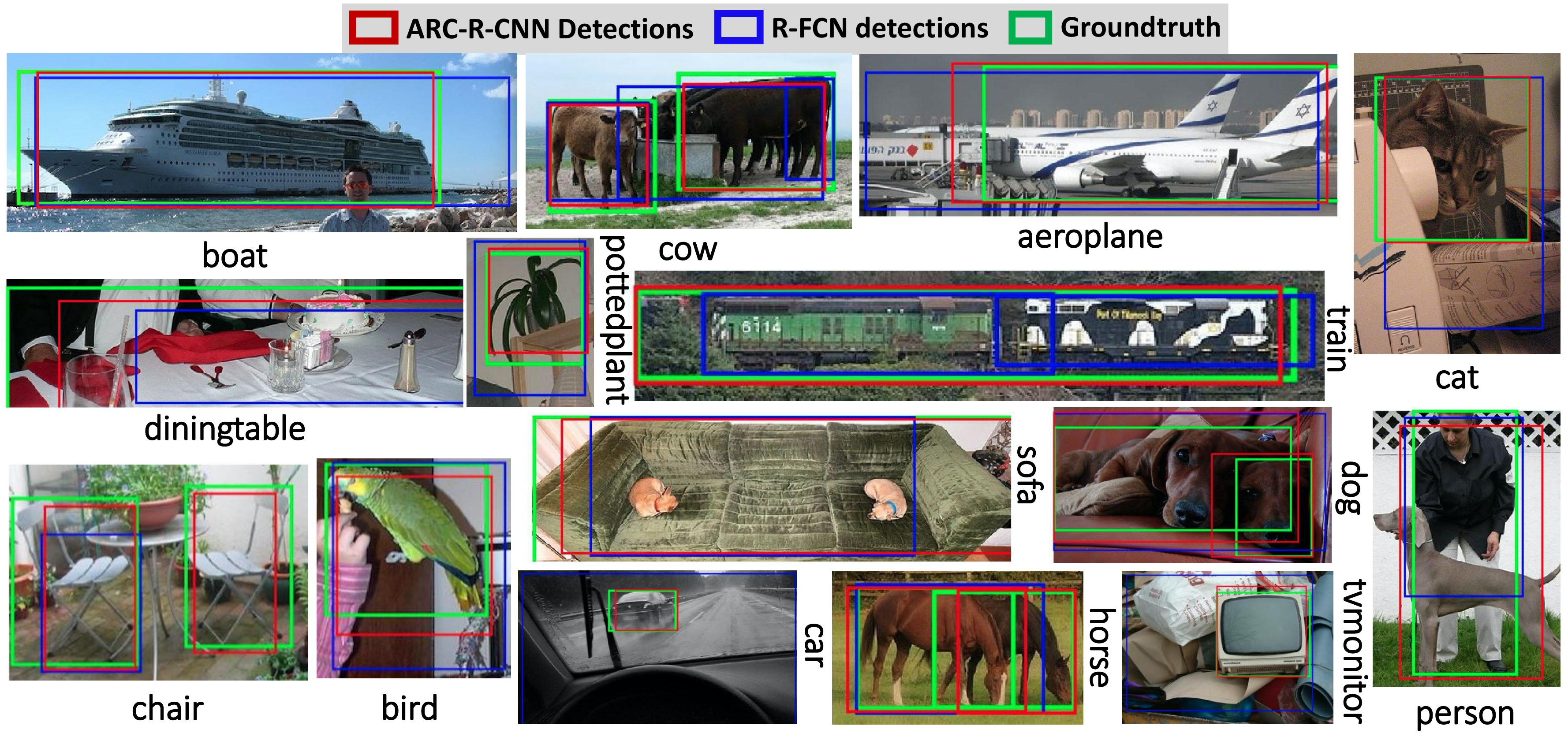}}
\caption{Sample detections of ARC-R-CNN-Res101 (red) and R-FCN-Res101 \cite{rfcn} (blue). For comparison, we also show the groundtruth bounding box (green). The score threshold is set to $0.6$ for good visualization. Best viewed in color and zoom in. \vspace{-5mm} }
\label{fig:comparisons}
\vspace{-1mm}
\end{figure*} 

\textbf{Accurate Object Localization on VOC 2007.} To verify our model on the ability of accurate localizations, we also utilize the IoU $\ge 0.7$ evaluation criterion.
Table \ref{tab:07fine} shows the results of our model and state-of-the-art methods, we can see ARC-R-CNN outperms R-FCN by a large margin $6.9$ and $7.7$ points in terms of mAP with ResNet-50 and ResNet-101, respectively. 
Comparing Table \ref{tab:07fine} with Table \ref{tab:07coarse}, we can see ARC-R-FCN enjoys more performance gains on more accurate localization (i.e., IoU $\ge 0.7$ vs. IoU $\ge 0.5$).
These results also show that bounding box aspect ratio and context are more important on accurately localizing the positions of objects with diverse shapes, which is the motivation of this work.

In Fig. \ref{fig:comparisons}, we show some qualitative results of ARC-R-CNN (red) and R-FCN (blue, using the ResNet-101 based model provided by the authors of \cite{rfcn}.). For the convenience of comparison, we also show the annotations (green). From these results, we can see our model can localize objects more accurately than the R-FCN baseline.

From Table \ref{tab:07fine}, we can see our model also outperforms previously state-of-the-art models that have better abilities on localizations than R-CNN style models \cite{rcnn,fast_rcnn,faster_rcnn}. In specific, ARC-R-CNN-Res50 is already comparative with the best LocNet in \cite{locNet}, and ARC-R-CNN-Res101 achieves $2.8$ points improvement over the best LocNet in \cite{locNet}.
Since \cite{yuting, locNet} using different strategies on predicting precise object positions, i.e, applying Bayesian optimization and structural prediction \cite{yuting}, and modelling localization probabilities \cite{locNet}, they are complementary with our method.

\begin{figure*}
\centering
{\includegraphics[width = 1.0\textwidth]{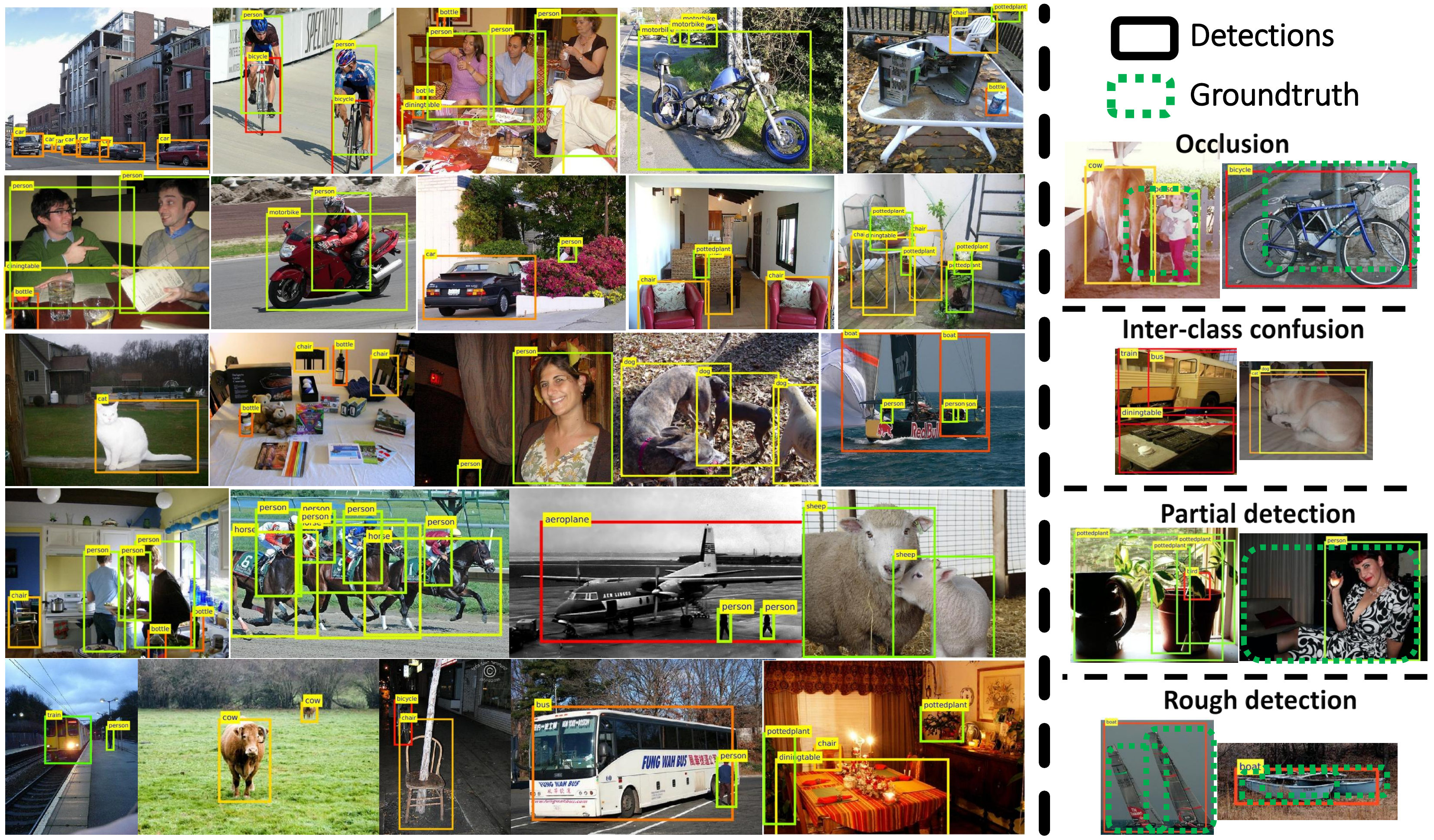}}
\caption{Qualitative Results of ARC-R-CNN-Res101 on Pascal VOC 2007. The score threshold is set to $0.45$ for good visualization. Best viewed in color and zoom in. \vspace{-4mm} }
\label{fig:dets}
\vspace{-1mm}
\end{figure*} 

\textbf{PASCAL VOC 2012 Testset}.
For VOC 2012 benchmark \cite{pascal}, we train on VOC 2007 \textit{trainval+test} and VOC 2012 \textit{trainval} (``07++12") following \cite{faster_rcnn,rfcn}, and evaluate on VOC 2012 \textit{test}. 
Training and testing strategies are the same as VOC 2007.
Table \ref{tab:12coarse} shows the IoU $\ge 0.5$ results of our method and state-of-the-art Faster-RCNN \cite{faster_rcnn,resNet} and R-FCN \cite{rfcn}.
For fair comparison, we reimplement R-FCN based on ResNet-101 without multi-scale training. 
We can see our method still outperforms them by $4.5$ and $1.7$ points of mAP respectively. 
The testing time of ARC-R-FCN and baseline models is also listed in Table \ref{tab:07coarse} and Table \ref{tab:12coarse}, we can see ARC-R-CNN doesn't increasing the test time much, just about $380ms$ per image, though slower than R-FCN, still faster than Faster RCNN.

\begin{table} 
\begin{center}
\resizebox{0.95\hsize}{!}{
\begin{tabular}{|c|c|c|c|c|}
\hline
\multicolumn{4}{|c|}{PASCAL VOC 2012 (IoU $\ge 0.5$)}  \\
\hline
Method & training data & test times & mAP\\
\hline
Faster-RCNN-Res101 &$07$++$12$ & $0.42$ & $73.8$  \\
R-FCN-Res101-ReIm$\dag:$ & $07$++$12$ & $0.17$ & $76.7$  \\
ARC-R-CNN-Res101$\ddag:$ & $07$++$12$ & $0.38$ & $\mathbf{78.4}$  \\
\hline
\end{tabular}
}
\end{center}
\caption{mAP results with IoU $\ge 0.5$ of Faster-RCNN, R-FCN and ARC-R-CNN on PASCAL VOC 2012 test set. Time is evaluated on a Nvidia K40 GPU. $\dag:$ \small{http://host.robots.ox.ac.uk:8080/anonymous/C3H0GM.html} $\ddag:$ \small{http://host.robots.ox.ac.uk:8080/anonymous/WB5KF0.html}}
\label{tab:12coarse} 
\vspace{-3mm}
\end{table}

Fig. \ref{fig:dets} shows some qualitative results of our model on the PASCAL VOC 2007 testset. For visual convenience, different object classes are shown with different colors. 
On the left, we show some detection examples, we set a relatively high threshold of $0.45$ for good displaying. From these results we can see, our model can fairly well localizing various objects. 
On the right, we show the typical failure examples. The failure cases are mainly due to occlusion, rough or partial detections and inter-class confusions (e.g., cat and dog, bus and train).

\vspace{-1mm}
\subsection{Preliminary MS COCO Results}
\vspace{-1mm}
To further validate our model, we train ARC-R-CNN with ResNet-101 as the backbone on the MS COCO dataset. The $80k$ \textit{train} set is utilized for training, and $40k$ \textit{val} set is utilized for testing. Multi-Stage Training is adopted. For Stage1, we use a learning rate $0.0005$ for the first $170k$ iterations and $0.00005$ for the next $60k$ iterations, with an effective mini-batch size of $8$. For Stage2, we set the learning rate of $0.0005$ for $80k$ iterations and $0.00005$ for next $30k$ iterations.

\begin{table} 
\begin{center}
\resizebox{1.0\hsize}{!}{
\begin{tabular}{c|c|c|c|c|c}
Method &  data &  data & AP@0.5 & AP@0.75 & AP\\
\hline
Faster-RCNN-Res101 & train & val & $48.4$ & - & $27.2$  \\
R-FCN-Res101 & train & val & $48.9$ & - & $27.6$  \\
py-R-FCN-Res101$\dag$ & train & val & $47.6$ & $29.3$ & $27.9$ \\
ARC-R-CNN-Res101 & train & val & $\mathbf{51.4}$ & $\mathbf{35.3}$ & $\mathbf{32.5}$  \\
\end{tabular}
}
\end{center}
\caption{Preliminary detection results on MS COCO. $\dag$ \small{https://github.com/Orpine/py-R-FCN}}
\label{tab:coco} 
\vspace{-3mm}
\end{table}

Table \ref{tab:coco} shows the preliminary results on COCO \textit{val} set, we can see our model outperforms Faster-RCNN and R-FCN on all AP@0.5, AP@0.75, and AP[0.5:0.95] evaluation protocols.
Specifically, ARC-R-CNN outperforms Fast R-CNN and R-FCN by $2.5 \sim 3.0$ points on AP@0.5 and $4.9 \sim 5.3$ points on AP@[0.5:0.95].
To validate the improvements over baseline model on AP@0.75, we compare our model with the python version R-FCN-Res101 provided by the authors of \cite{rfcn}.
we can see ARC-R-CNN-Res101 outperforms py-R-FCN-Res101 by $6.0$ points on AP@0.75, this is similar to the observations on Pascal VOC 2007 dataset.

\vspace{-1mm}
\subsection{Ablation Study}
\begin{figure}
\centering
{\includegraphics[width = 0.45\textwidth]{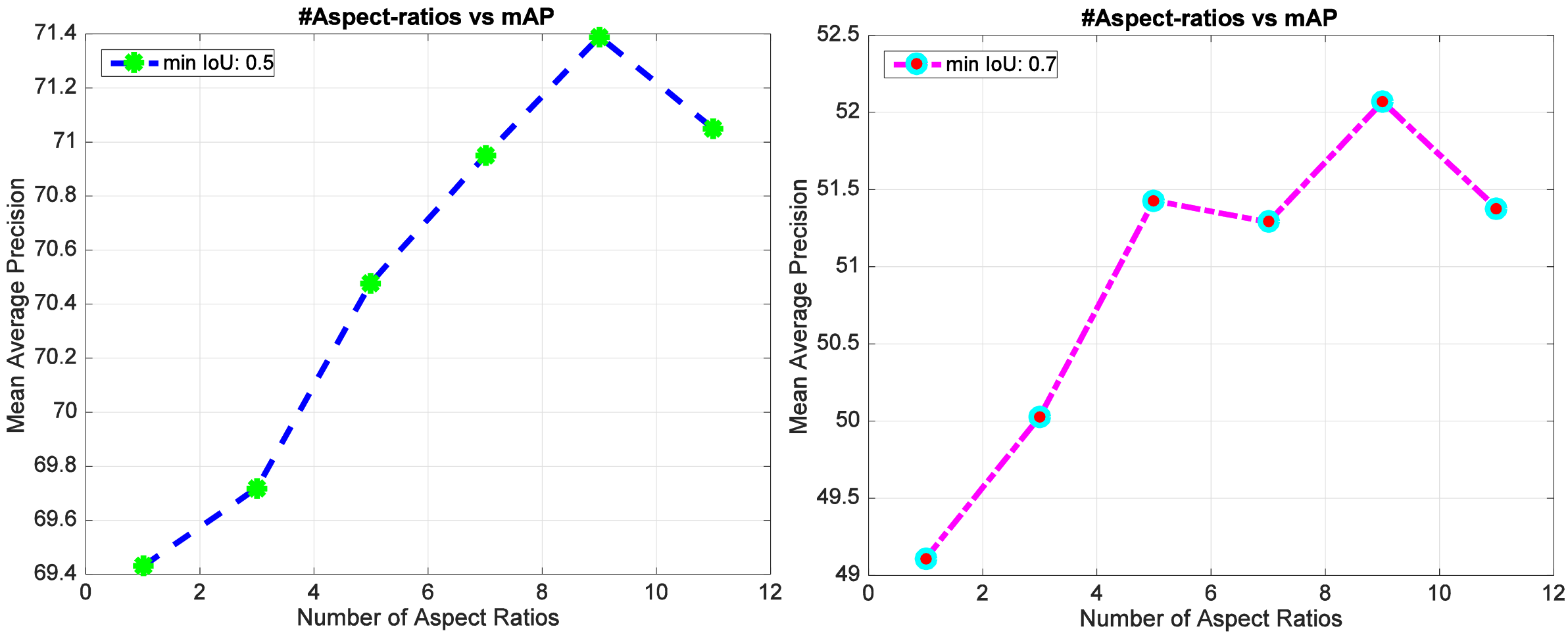}}
\caption{mAP results with different number of aspect ratio branches on PASCAL VOC 2007 test set. \vspace{-3mm} }
\label{fig:aspects}
\end{figure}

\vspace{-1mm}
\textit{Effects of multi-aspect ratios}.
To investigate the role of aspect ratio modelling, we trained several ARC-R-CNNs but without context modelling, these models have different number of aspect ratios. 
We utilize the ResNet-50 for initialization and set the learning rate as $0.001$ for the first $50$k iterations training and $0.0001$ for the next $20$k iterations, with an effective mini-batch size of $1$. For this pilot experiment, OHEM is not used.
We investigate aspect ratios with five branches: a) $\{7\times7\}$, b) $\{7\times7, 5\times10, 10\times5\}$, c) $\{7\times7, 5\times10, 10\times5, 3\times12, 12\times3\}$, d) $\{7\times7, 5\times10, 10\times5, 4\times12, 12\times4,3\times12, 12\times3\}$, e)$\{7\times7, 5\times10, 10\times5, 4\times12, 12\times4,3\times12, 12\times3, 3\times15, 15\times3\}$.
Fig. \ref{fig:aspects} shows the results, we can see increasing the number of aspect ratios will also increase the detection performance in general.
However, when introducing too many asspect ratio branches, the performance may drops due to overfitting. This may be relieved by introducing more data or some effective regularization techniques (e.g., dropout \cite{dropout}).

\textit{Effects of context}.
To investigate the impact of context, we trained several ARC-R-CNNs but without aspect ratio modelling.
For this experiment, we utilize the ResNet-101 for weight initialization and set the learning rate as $0.001$ for the first $80$k iterations training, and $0.0001$ for the next $30$k iterations, with an effective mini-batch size of $2$. OHEM is adopt for training. 
Table \ref{tab:context} shows the detailed results of ARC-R-CNN with/without local and global contexts, we can see both local context and global context can boost the detection performance. Local context is not used for bounding box regression in ARC-R-CNN-local-global-var1, but used in ARC-R-CNN-local-global-var2.
We can see local context can help localizing objects when min IoU is $0.5$, but harms the result when the min IoU is $0.7$.
In our preliminary experiment (with ResNet-50), we find the global context harms the bounding box regression by $2.9$ points, so we don't  explore global context for bounding box regression here.

\begin{table} 
\begin{center}
\resizebox{0.85\hsize}{!}{
\begin{tabular}{|c|c|c|c|c|c|c|}
\hline
\multicolumn{3}{|c|}{Local \& Global Context (PASCAL VOC 2007)}  \\
\hline
Method & IoU $\ge 0.5$ & IoU $\ge 0.7$ \\
\hline
ARC-R-CNN-no-context & $79.4$ & $60.5$  \\
\hline
ARC-R-CNN-global & $79.9$ & $60.6$ \\
\hline
ARC-R-CNN-local-global-var1 & $80.4$ & $63.3$ \\
\hline
ARC-R-CNN-local-global-var2 & $\mathbf{80.7}$ & $62.8$ \\
\hline
\end{tabular}
}
\end{center}
\caption{mAP results with local and global context on PASCAL VOC 2007 test set. Local context is not used for bounding box regression in ARC-R-CNN-Res101-local-global-context$^1$, but used in ARC-R-CNN-Res101-local-global-context$^2$.}
\label{tab:context} 
\vspace{-2mm}
\end{table}

\textit{Effects of Multi-Stage ARC-R-CNN Training}. \label{sec:2step}
We compare ARC-R-CNNs trained with different iterations on the Pascal VOC and MS COCO datasets. 
To analyze the effect of each training step, we compare the results of both IoU $\ge 0.5$ and IoU $\ge 0.7/0.75$.
Table~\ref{tab:two_step} show the results.
On VOC 2007 dataset, we can see i) the Stage1 training already gets powerful results on both IoU $\ge 0.5$ and IoU $\ge 0.7$, ii) the Stage2 training doesn't improve the mAP for IoU $\ge 0.5$,
 but improves the mAP by $3.5 \sim 3.8$ points for IoU $\ge 0.7$.
On MS COCO dataset, both the Stage1 and Stage2 training increase the detection results, and the gain for IoU $\ge 0.75$ is bigger than that for IoU $\ge 0.5$.
From this experiment, we can see multi-stage training is very useful for accurate localization.

\begin{table} 
\begin{center}
\resizebox{0.8\hsize}{!}{
\begin{tabular}{|c|c|c|c|c|}
\hline
\multicolumn{3}{|c|}{Multi-Stage Training (PASCAL VOC 2007)}  \\
\hline
Method &  IoU $\ge 0.5$ & IoU $\ge 0.7$ \\
\hline
R-FCN-Res50-ReIm & $77.3$ & $57.8$  \\
ARC-R-CNN-Res50-Stage1 & $\textbf{80.2}$  & $60.2$ \\
ARC-R-CNN-Res50-Stage2 & $80.1$ & $\textbf{64.0}$ \\
\hline
R-FCN-Res101-ReIm & $79.4$ & $60.5$  \\
ARC-R-CNN-Res101-Stage1 & $\mathbf{82.0}$ & $64.7$ \\
ARC-R-CNN-Res101-Stage2 & $81.7$ & $\mathbf{68.2}$  \\
\hline
\multicolumn{3}{|c|}{Multi-Stage Training (MS COCO val)}  \\
\hline
Method &  IoU $\ge 0.5$ & IoU $\ge 0.75$ \\
\hline
ARC-R-CNN-Res101-Stage1 & $50.2$ & $32.5$  \\
ARC-R-CNN-Res101-Stage2 & $\mathbf{51.4}$ & $\mathbf{35.3}$ \\
\hline
\end{tabular}
}
\end{center}
\caption{Comparison of the Stage1 and Stage2 ARC-R-CNNs on mAP with both IoU $\ge 0.5$ and IoU $\ge 0.7/0.75$ on PASCAL VOC 2007 and MS COCO datasets.}
\label{tab:two_step}
\vspace{-2mm}
\end{table}

\vspace{-2mm}
\section{Conclusion}
\vspace{-1mm}
This paper presented the aspect ratio and context aware  region-based convolutional neural network (ARC-RCNN), which integrates aspect ratio and multi-scale context in R-CNN in an end-to-end way for accurate object detection. The key idea is to enrich the region-of-interest (RoI) pooling widely used in R-CNN based object detection systems.  Unlike existing R-CNN based methods which utilize a single type of tiling in the RoI pooling (e.g., with equally-sized $7\times 7$ cells), ARC-R-CNN exploits a mixture of tiling configurations, which not only remedies the warping artifacts caused by a single RoI pooling, but also respects the underlying object shapes more. ARC-R-CNN also ``looks from outside of the RoI" by incorporating contextual information at two scales: global context pooled from the whole image  and local context pooled from the surrounding of a RoI, both with the proposed mixture of tiling configurations. 
Besides, we also propose a multi-stage training framework which endows ARC-R-CNN powerful abilities on accurate object detection.
ARC-R-CNN outperforms both Faster R-CNN~\cite{faster_rcnn} and R-FCN~\cite{rfcn} with significantly better average precision using larger value for IoU $\ge 0.7$.

\begin{table*} 
\begin{center}
\resizebox{1.0\hsize}{!}{
\begin{tabular}{|c|c|c|c|c|c|c|c|c|c|c|c|c|c|c|c|c|c|c|c|c|c|c|}
\hline
method & areo & bike & bird & boat & bottle & bus & car & cat & chair & cow & table & dog & horse & mbike & person & plant & sheep & sofa & train & tv & mAP \\
\hline
Faster-RCNN-Res101 & $79.8$ & $80.7$ & $76.2$ & $68.3$ & $55.9$ & $85.1$ & $85.3$ & $89.8$ & $56.7$ & $87.8$ & $69.4$ & $88.3$ & $88.9$ & $80.9$ & $78.4$ & $41.7$ & $78.6$ & $79.8$ & $85.3$ & $72.0$ & $76.4$ \\
R-FCN-Res101 & $82.5$ & $83.7$ & $80.3$ & $69.0$ & $69.2$ & $87.5$ & $88.4$ & $88.4$ & $65.4$ & $87.3$ & $72.1$ & $87.9$ & $88.3$ & $81.3$ & $79.8$ & $54.1$ & $79.6$ & $78.8$ & $87.1$ & $79.5$ & $79.5$ \\
ARC-FCN-Res101 & $\mathbf{88.3}$ & $\mathbf{88.0}$ & $\mathbf{83.4}$ & $\mathbf{75.8}$ & $\mathbf{71.0}$ & $\mathbf{88.0}$ & $\mathbf{89.2}$ & $\mathbf{89.6}$ & $\mathbf{68.5}$ & $\mathbf{88.3}$ & $\mathbf{77.3}$ & $\mathbf{88.9}$ & $\mathbf{88.7}$ & $\mathbf{85.1}$ & $\mathbf{83.2}$ & $\mathbf{54.3}$ & $\mathbf{83.7}$ & $\mathbf{80.6}$ & $\mathbf{87.7}$ & $\mathbf{80.7}$ & $\mathbf{82.0}$  \\
\hline
\end{tabular}
}
\end{center}
\caption{Detailed results with IoU $\ge 0.5$ of Faster-RCNN, R-FCN and ARC-FCN on PASCAL VOC 2007 test set.}
\label{tab:07coarse-supp} 
\end{table*}

\begin{table*} 
\begin{center}
\resizebox{1.0\hsize}{!}{
\begin{tabular}{|c|c|c|c|c|c|c|c|c|c|c|c|c|c|c|c|c|c|c|c|c|c|c|}
\hline
method & areo & bike & bird & boat & bottle & bus & car & cat & chair & cow & table & dog & horse & mbike & person & plant & sheep & sofa & train & tv & mAP \\
\hline
R-FCN-Res101 & 61.7 & 66.9 & 58.0 & 43.4 & 41.5 & 78.9 & 70.7 & 74.4 & 41.0 & 68.6 & 47.1 & 72.3 & 69.6 & 62.9 & 57.1 & 31.2 & 67.0 & 60.1 & 70.5 & 66.9 & 60.5 \\
ARC-FCN-Res101 & $\mathbf{78.1}$ & $\mathbf{76.3}$ & $\mathbf{66.6}$ & $\mathbf{54.5}$ & $\mathbf{58.2}$ & $\mathbf{80.4}$ & $\mathbf{78.8}$ & $\mathbf{78.2}$ & $\mathbf{51.0}$ & $\mathbf{70.9}$ & $\mathbf{55.5}$ & $\mathbf{74.1}$ & $\mathbf{75.9}$ & $\mathbf{69.0}$ & $\mathbf{67.1}$ & $\mathbf{39.5}$ & $\mathbf{74.8}$ & $\mathbf{64.2}$ & $\mathbf{76.7}$ & $\mathbf{74.2}$ & $\mathbf{68.2}$  \\
\hline
\end{tabular}
}
\end{center}
\caption{Detailed results with IoU $\ge 0.7$ of R-FCN and ARC-FCN on PASCAL VOC 2007 test set.}
\label{tab:07fine-supp} 
\end{table*}

\begin{table*}
\begin{center}
\resizebox{1.0\hsize}{!}{
\begin{tabular}{|c|c|c|c|c|c|c|c|c|c|c|c|c|c|c|c|c|c|c|c|c|c|c|}
\hline
method & areo & bike & bird & boat & bottle & bus & car & cat & chair & cow & table & dog & horse & mbike & person & plant & sheep & sofa & train & tv & mAP \\
\hline
Faster-RCNN-Res101 & 86.5 & 81.6 & 77.2 & 58.0 & 51.0 & 78.6 & 76.6 & 93.2 & 48.6 & 80.4 & 59.0 & 92.1 & 85.3 & 84.8 & 80.7 & 48.1 & 77.3 & 66.5 & 84.7 & 65.6 & 73.8\\
R-FCN-Res101-ReIm$\dag$ & $86.9$ & $84.0$ & $78.4$ & $63.6$ & $59.6$ & $80.5$ & $80.5$ & $91.8$ & $58.6$ & $81.6$ & $58.8$ & $91.1$ & $85.0$ & $85.7$ & $83.7$ & $57.8$ & $81.0$ & $66.3$ & $85.1$ & $72.9$ & $76.7$ \\
ARC-FCN-Res101$\ddag$ & $\mathbf{88.7}$ & $\mathbf{84.3}$ & $\mathbf{81.7}$ & $\mathbf{66.4}$ & $\mathbf{63.0}$ & $\mathbf{81.6}$ & $\mathbf{82.1}$ & $\mathbf{92.8}$ & $\mathbf{60.2}$ & $\mathbf{84.4}$ & $\mathbf{60.6}$ & $\mathbf{92.3}$ & $\mathbf{87.8}$ & $\mathbf{86.4}$ & $\mathbf{85.3}$ & $\mathbf{60.2}$ & $\mathbf{82.0}$ & $\mathbf{66.6}$ & $\mathbf{86.9}$ & $\mathbf{74.4}$ & $\mathbf{78.4}$  \\
\hline
\end{tabular}
}
\end{center}
\caption{Detailed results with IoU $\ge 0.5$ of Faster-RCNN, R-FCN and ARC-FCN on PASCAL VOC 2012 test set. $\dag:$ \small{http://host.robots.ox.ac.uk:8080/anonymous/C3H0GM.html}. $\ddag:$ \small{http://host.robots.ox.ac.uk:8080/anonymous/WB5KF0.html}
}
\label{tab:12coarse-supp} 
\end{table*}

\begin{appendix}
\section{Detailed Results on PASCAL VOC Datasets}
In the article, we compared with Faster R-CNN \cite{faster_rcnn} and R-FCN \cite{rfcn}, with all methods using their pure forms (no multi-scale train/test and other tricks), to focus on the investigation of aspect ratio and context. We show the detailed results on PASCAL VOC Datasets \cite{pascal} in the following. 

\subsection{Pascal VOC 2007 Test Set.} 
In this section, we  show the detailed detection results of Faster R-CNN \cite{faster_rcnn}, R-FCN \cite{rfcn}, and ARC-FCN on PASCAL VOC 2007 test set. 
For this dataset, the union set of PASCAL VOC 2007 \textit{trainval} and PASCAL VOC 2012 \textit{trainval} is used as the training set, and the PASCAL VOC 2007 \textit{test} is used as the test set following \cite{faster_rcnn,rfcn}.
We adopt the PASCAL VOC evaluation protocol \cite{pascal}, and evaluate our model by mean average precision (mAP).
Both intersection over union (IoU) $\ge 0.5$ and $\ge 0.7$ are utilized for PASCAL VOC 2007 (not apply for PASCAL VOC 2012), as detailed annotations are available only
on PASCAL VOC 2007.

Table \ref{tab:07coarse-supp} shows the results with $IoU \ge 0.5$. Here, ARC-FCN is trained with One-Step training (excluding RPN training). We can see ARC-FCN outperforms the other $2$ methods on all object categories. For some object classes, the improvement is quite impressive (above $3$ points in terms of AP), e.g., areoplane, bike, bird, boat, chair, table, and person.

Table \ref{tab:07fine-supp} shows the results with $IoU \ge 0.7$. Here, ARC-FCN is trained with Two-Step training (excluding RPN training). The model of R-FCN-Res101 is provided by the authors of \cite{rfcn}. As R-FCN is superior than Faster R-CNN, we just compare our method with R-FCN here. 
We can see ARC-RFCN outperforms R-FCN by a even larger margin across all object categories. For some object classes, e.g., aeroplane, bike, boat, bottle, chair, person, the improvements is above $10$ points in terms of AP. These results verified the superior localization ability of ARC-FCN.

\begin{figure*}
\centering
{\includegraphics[width = 0.8\textwidth]{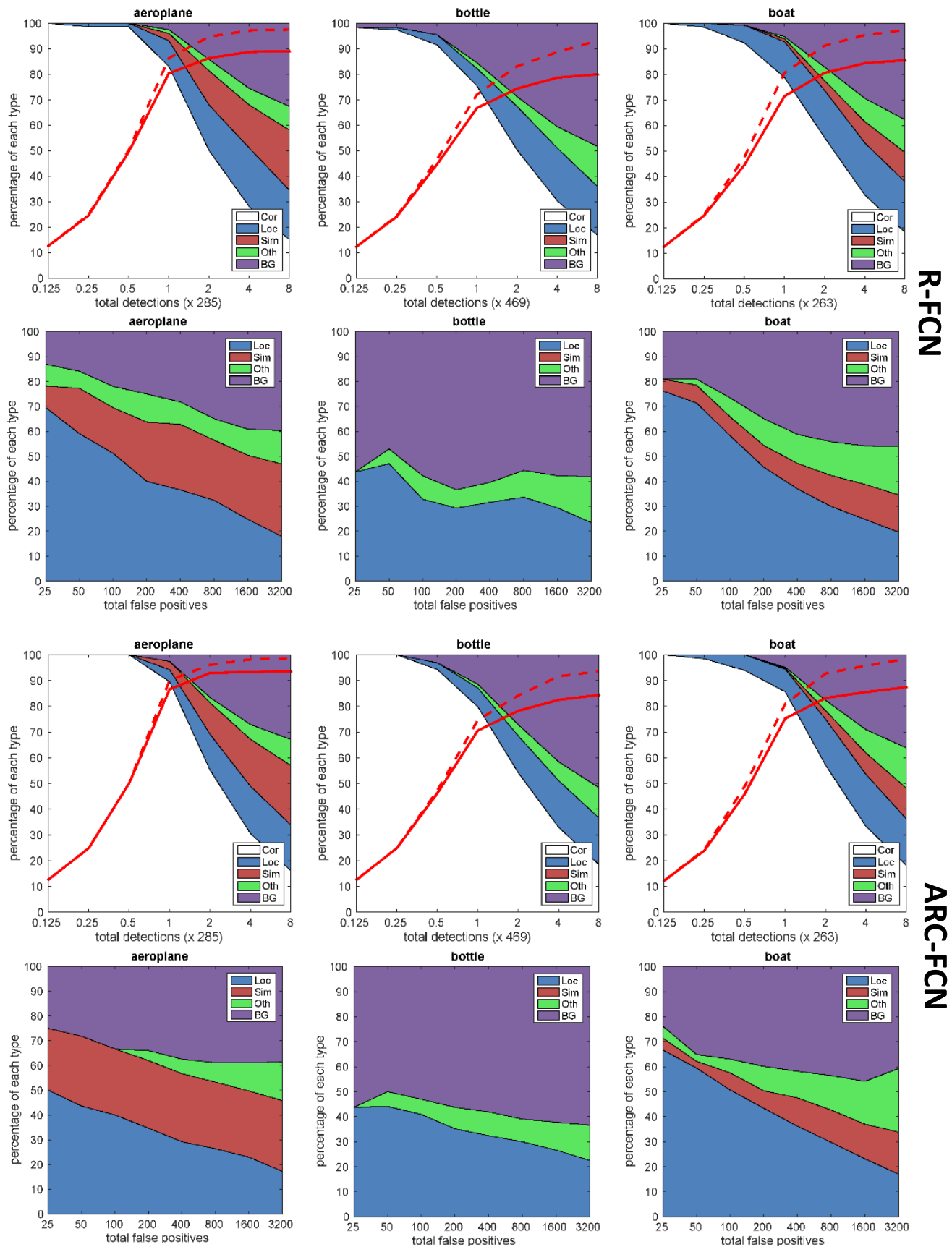}}
\caption{False positive/detection trends with rank. 
Detection analysis of R-FCN are shown on the first $2$ rows. Detection analysis of ARC-FCN are shown on the last $2$ rows.
The First and Third Rows: type of detection as number of detections increases; 
The Second and Fourth Rows: stacked area plot showing fraction of FP of each type as the total number of FP increase.
line plots show recall as function of the number of objects (dashed=weak localization, solid=strong localization).}
\label{fig:diag}
\end{figure*}

Besides, we also compare the detection diagnosis of R-FCN-Res101 and ARC-FCN-Res101 in Fig. \ref{fig:diag}. This is done by applying the excellent detection analysis tool from \cite{hoiem_diag}. 
For good visualization, we just show some typical object classes (i.e., aeroplane, bottle, boat) with large localization improvements by ARC-FCN.
The first two rows show the result of R-FCN, and the last two rows show the result of ARC-FCN. From these plots, we can see the localization error of ARC-FCN is smaller than the one of R-FCN.

\subsection{Pascal VOC 2012 Test Set.}
In this section, we show the detailed detection results ($IoU \ge 0.5$) of Faster R-CNN \cite{faster_rcnn}, R-FCN \cite{rfcn}, and ARC-FCN on PASCAL VOC 2012 dataset (see Table \ref{tab:12coarse-supp}). 
For this dataset, the union set of PASCAL VOC 2007 \textit{trainval+test} and PASCAL VOC 2012 \textit{trainval} is used as the training set, and the PASCAL VOC 2012 \textit{test} is used as the test set following \cite{faster_rcnn,rfcn}.
From Table \ref{tab:12coarse-supp}, we can also see ARC-FCN still outperforms Faster-RCNN and R-FCN across all object categories.

\section{More Qualitative Examples}

In addition to the article, we show more qualitative examples in this section.

\begin{figure*}
\centering
{\includegraphics[width = 0.96\textwidth]{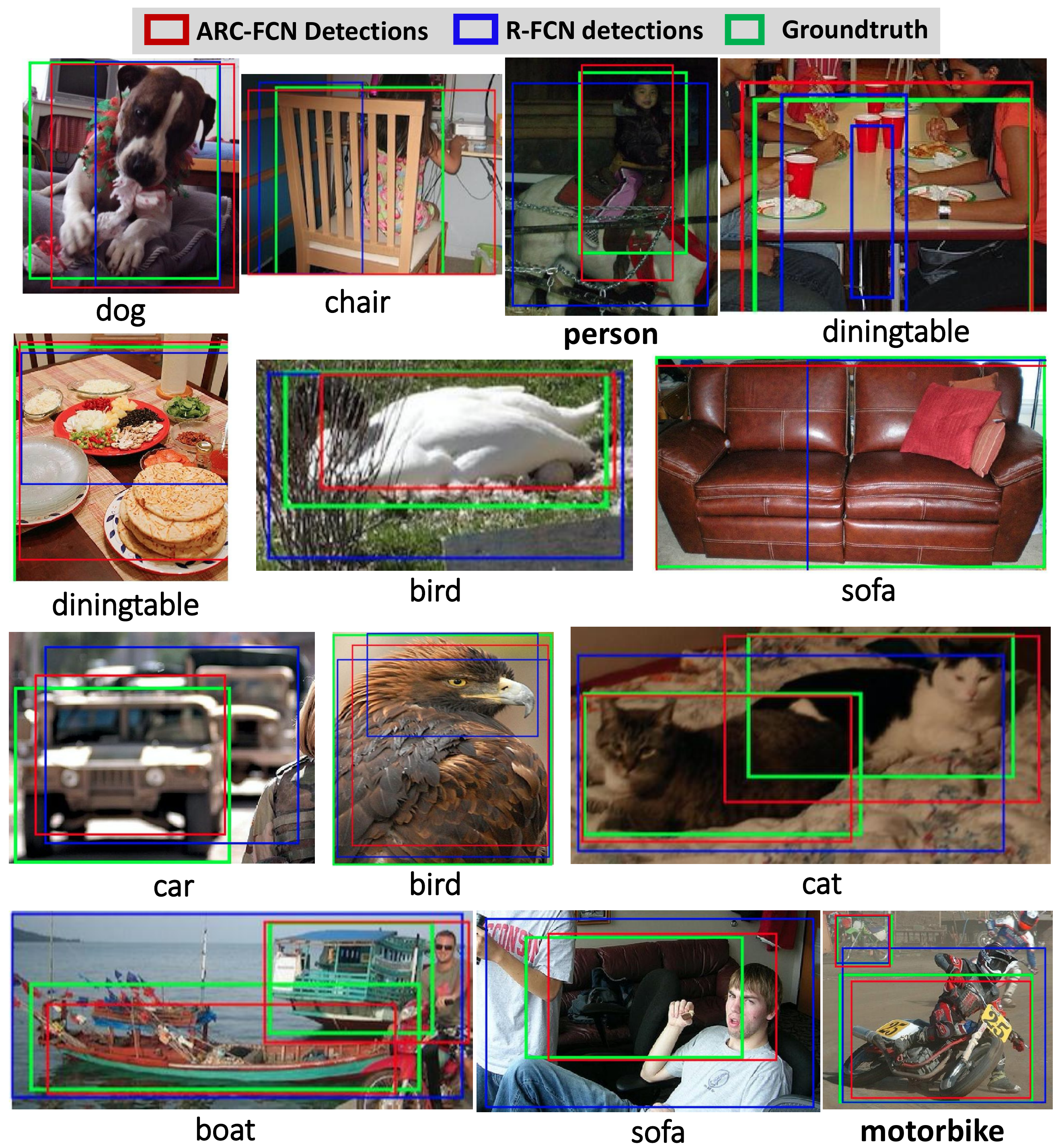}}
\caption{Sample detections of R-FCN-Res101 \cite{rfcn} (blue) and ARC-FCN-Res101 (red). For comparison, we also show the groundtruth bounding box (green). The score threshold is set to $0.6$ for good visualization. Best viewed in color and zoom in. 
}
\label{fig:comp_supp}
\end{figure*} 

\begin{figure*}
\centering
{\includegraphics[width = 0.96\textwidth]{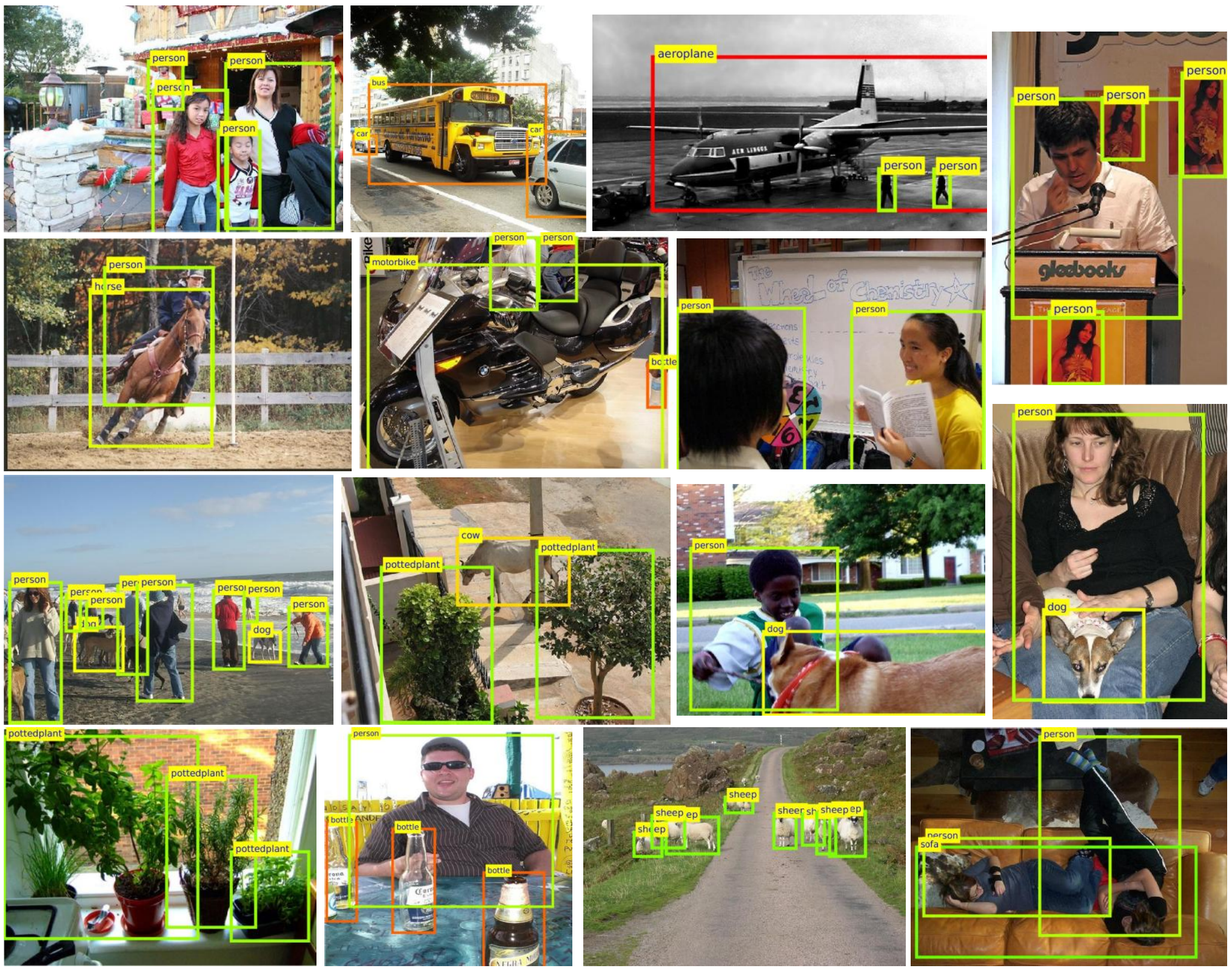}}
\caption{Qualitative Results of ARC-FCN-Res101 on Pascal VOC 2007 test set. For good displaying, different object classes are shown with different colors. The score threshold is set to $0.45$ for good visualization. Best viewed in color and zoom in. }
\label{fig:dets_supp}
\end{figure*} 

\begin{figure*}
\centering
{\includegraphics[width = 0.96\textwidth]{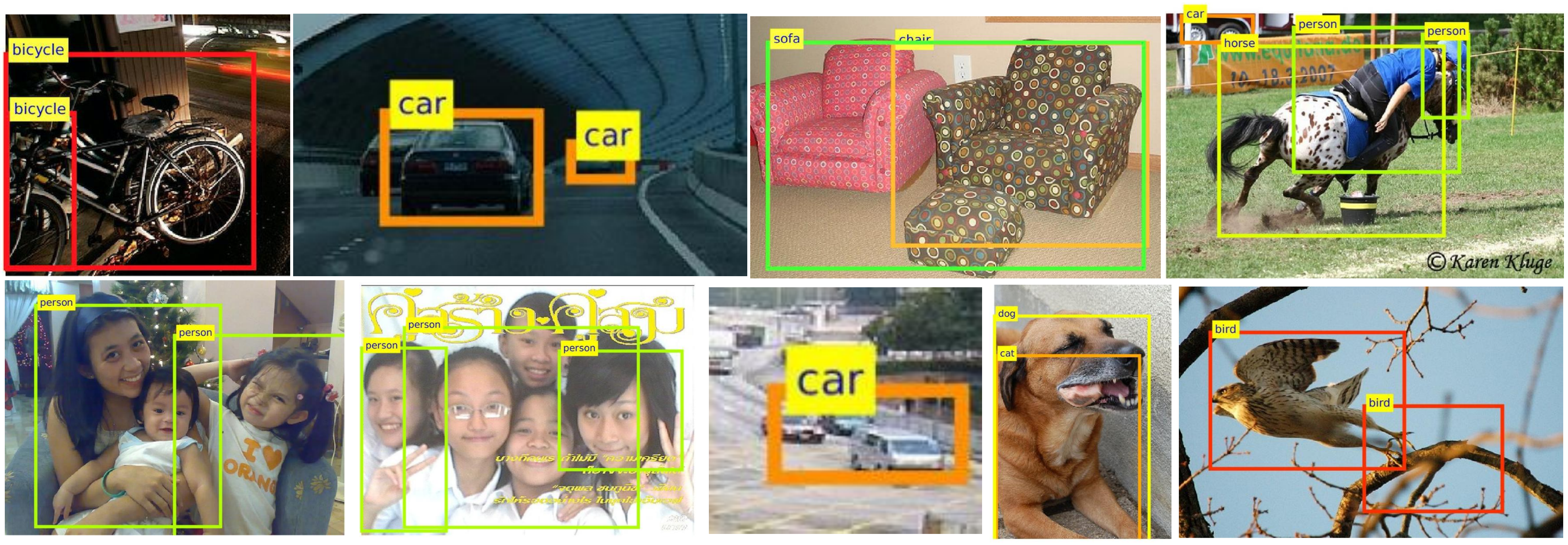}}
\caption{Failure examples of ARC-FCN-Res101 on Pascal VOC 2007 test set. 
Different object classes are shown with different colors.
The score threshold is set to $0.45$ for good visualization.
The failure cases are mainly due to occlusion, rough or partial detections, and inter-class or foreground/background confusions.
Best viewed in color and zoom in. }
\label{fig:failure_supp}
\end{figure*}

Fig. \ref{fig:comp_supp} shows some qualitative results of ARC-FCN (red) and R-FCN (blue). For the convenience of comparison, the groundtruth annotations are also showed by the green bounding boxes. From these results, we can see our model can localize objects more accurately than R-FCN.

Fig. \ref{fig:dets_supp} shows some qualitative results of our model on the PASCAL VOC 2007 testset. 
We set a relatively high threshold of $0.45$ for good visualization, and different object classes are shown with different colors. From these results, we can see our model can fairly well localizing various objects with diverse shapes and imaging conditions.

Fig. \ref{fig:failure_supp} shows some typical failure examples of our model. The failure cases are mainly due to occlusion, rough or partial detections and inter-class confusions (e.g., cat and dog, sofa and chair).
From the last figure in Fig. \ref{fig:failure_supp} and the detection analysis in Fig. \ref{fig:diag}, we can also see another failure case is the confusion of objects and various backgrounds.
\end{appendix}

{\small
\bibliographystyle{ieee}
\bibliography{arc_fcn}

\begin{thebibliography}{10}\itemsep=-1pt

\bibitem{mixture_dpm}
O.~Aghazadeh, H.~Azizpour, J.~Sullivan, and S.~Carlsson.
\newblock Mixture component identification and learning for visual recognition.
\newblock In A.~Fitzgibbon, S.~Lazebnik, P.~Perona, Y.~Sato, and C.~Schmid,
  editors, {\em ECCV}, 2012.

\bibitem{ssdpm}
H.~Azizpour and I.~Laptev.
\newblock Object detection using strongly-supervised deformable part models.
\newblock In {\em ECCV}, 2012.

\bibitem{ion}
S.~Bell, C.~L. Zitnick, K.~Bala, and R.~B. Girshick.
\newblock Inside-outside net: Detecting objects in context with skip pooling
  and recurrent neural networks.
\newblock In {\em CVPR}, 2016.

\bibitem{cao2016realtime}
Z.~Cao, T.~Simon, S.-E. Wei, and Y.~Sheikh.
\newblock Realtime multi-person 2d pose estimation using part affinity fields.
\newblock {\em arXiv preprint arXiv:1611.08050}, 2016.

\bibitem{hole}
L.-C. Chen, G.~Papandreou, I.~Kokkinos, K.~Murphy, and A.~L. Yuille.
\newblock Semantic image segmentation with deep convolutional nets and fully
  connected crfs.
\newblock In {\em ICLR}, 2015.

\bibitem{mono3d}
X.~Chen, K.~Kundu, Z.~Zhang, H.~Ma, S.~Fidler, and R.~Urtasun.
\newblock Monocular 3d object detection for autonomous driving.
\newblock In {\em CVPR}, 2016.

\bibitem{xianjie}
X.~Chen and A.~Yuille.
\newblock Articulated pose estimation by a graphical model with image dependent
  pairwise relations.
\newblock In {\em NIPS}, 2014.

\bibitem{leozhu_nips07}
Y.~Chen, L.~Zhu, C.~Lin, A.~L. Yuille, and H.~Zhang.
\newblock Rapid inference on a novel and/or graph for object detection,
  segmentation and parsing.
\newblock In {\em NIPS}, 2007.

\bibitem{BING}
M.-M. Cheng, Z.~Zhang, W.-Y. Lin, and P.~H.~S. Torr.
\newblock {BING}: Binarized normed gradients for objectness estimation at
  300fps.
\newblock In {\em IEEE CVPR}, 2014.

\bibitem{rfcn}
J.~Dai, Y.~Li, K.~He, and J.~Sun.
\newblock {R-FCN}: Object detection via region-based fully convolutional
  networks.
\newblock {\em arXiv preprint arXiv:1605.06409}, 2016.

\bibitem{hoiem_loc}
Q.~Dai and D.~Hoiem.
\newblock Learning to localize detected objects.
\newblock In {\em Computer Vision and Pattern Recognition (CVPR), 2012 IEEE
  Conference on}, 2012.

\bibitem{desai}
C.~Desai, D.~Ramanan, and C.~C. Fowlkes.
\newblock Discriminative models for multi-class object layout.
\newblock {\em International Journal of Computer Vision}, 2011.

\bibitem{szegedy_cvpr14}
D.~Erhan, C.~Szegedy, A.~Toshev, and D.~Anguelov.
\newblock Scalable object detection using deep neural networks.
\newblock In {\em CVPR}, 2014.

\bibitem{pascal}
M.~Everingham, L.~Van~Gool, C.~Williams, J.~Winn, and A.~Zisserman.
\newblock The pascal visual object classes (voc) challenge.
\newblock {\em IJCV}, 2010.

\bibitem{DPM}
P.~Felzenszwalb, R.~Girshick, D.~McAllester, and D.~Ramanan.
\newblock Object detection with discriminatively trained part-based models.
\newblock {\em PAMI}, 2010.

\bibitem{fidler}
S.~Fidler, R.~Mottaghi, A.~Yuille, and R.~Urtasun.
\newblock Bottom-up segmentation for top-down detection.
\newblock In {\em CVPR}, 2013.

\bibitem{mrcnn}
S.~Gidaris and N.~Komodakis.
\newblock Object detection via a multi-region {\&} semantic segmentation-aware
  {CNN} model.
\newblock {\em CoRR}, abs/1505.01749, 2015.

\bibitem{locNet}
S.~Gidaris and N.~Komodakis.
\newblock Locnet: Improving localization accuracy for object detection.
\newblock In {\em CVPR}, 2016.

\bibitem{fast_rcnn}
R.~Girshick.
\newblock Fast r-cnn.
\newblock In {\em International Conference on Computer Vision ({ICCV})}, 2015.

\bibitem{rcnn}
R.~Girshick, J.~Donahue, T.~Darrell, and J.~Malik.
\newblock Rich feature hierarchies for accurate object detection and semantic
  segmentation.
\newblock In {\em CVPR}, 2014.

\bibitem{pff_grammar}
R.~Girshick, P.~Felzenszwalb, and D.~McAllester.
\newblock Object detection with grammar models.
\newblock In {\em NIPS}, 2011.

\bibitem{dpdpm}
R.~Girshick, F.~Iandola, T.~Darrell, and J.~Malik.
\newblock Deformable part models are convolutional neural networks.
\newblock In {\em CVPR}, 2015.

\bibitem{guangchen_cvpr}
J.~X. Guang~Chen, Yuanyuan~Ding and T.~X. Han.
\newblock Detection evolution with multi-order contextual co-occurrence.
\newblock In {\em CVPR}, 2013.

\bibitem{XAI}
D.~Gunning.
\newblock Explainable artificial intelligence, 2016.
\newblock
  \url{http://www.darpa.mil/program/explainable-artificial-intelligence}.

\bibitem{resNet}
K.~He, X.~Zhang, S.~Ren, and J.~Sun.
\newblock Deep residual learning for image recognition.
\newblock {\em arXiv preprint arXiv:1512.03385}, 2015.

\bibitem{hoiem_diag}
D.~Hoiem, Y.~Chodpathumwan, and Q.~Dai.
\newblock Diagnosing error in object detectors.
\newblock In {\em ECCV}, 2012.

\bibitem{hoiem06}
D.~Hoiem, A.~A. Efros, and M.~Hebert.
\newblock {Putting Objects in Perspective}.
\newblock In {\em IEEE CVPR}, 2006.

\bibitem{caffe}
Y.~Jia, E.~Shelhamer, J.~Donahue, S.~Karayev, J.~Long, R.~Girshick,
  S.~Guadarrama, and T.~Darrell.
\newblock Caffe: Convolutional architecture for fast feature embedding.
\newblock {\em arXiv preprint arXiv:1408.5093}, 2014.

\bibitem{sppnet}
H.~Kaiming, Z.~Xiangyu, R.~Shaoqing, and J.~Sun.
\newblock Spatial pyramid pooling in deep convolutional networks for visual
  recognition.
\newblock In {\em European Conference on Computer Vision}, 2014.

\bibitem{lecun89}
Y.~LeCun, B.~Boser, J.~S. Denker, D.~Henderson, R.~E. Howard, W.~Hubbard, and
  L.~D. Jackel.
\newblock Backpropagation applied to handwritten zip code recognition.
\newblock {\em Neural Computation}, 1:541--551, 1989.

\bibitem{carAOG}
B.~Li, T.~Wu, and S.-C. Zhu.
\newblock Integrating context and occlusion for car detection by hierarchical
  and-or model.
\newblock In {\em ECCV}, 2014.

\bibitem{haoxiang}
H.~Li, J.~Brandt, Z.~Lin, X.~Shen, and G.~Hua.
\newblock A multi-level contextual model for person recognition in photo
  albums.
\newblock In {\em The IEEE Conference on Computer Vision and Pattern
  Recognition (CVPR)}, June 2016.

\bibitem{coco}
T.~Lin, M.~Maire, S.~J. Belongie, L.~D. Bourdev, R.~B. Girshick, J.~Hays,
  P.~Perona, D.~Ramanan, P.~Doll{\'{a}}r, and C.~L. Zitnick.
\newblock Microsoft {COCO:} common objects in context.
\newblock {\em CoRR}, abs/1405.0312, 2014.

\bibitem{ssd}
W.~Liu, D.~Anguelov, D.~Erhan, C.~Szegedy, S.~Reed, C.-Y. Fu, and A.~C. Berg.
\newblock {SSD}: Single shot multibox detector.
\newblock {\em arXiv preprint arXiv:1512.02325}, 2015.

\bibitem{Lopez2011}
R.~J. Lopez-Sastre, T.~T., and S.~Savarese.
\newblock Deformable part models revisited: A performance evaluation for object
  category pose estimation.
\newblock In {\em ICCV-WS CORP}, 2011.

\bibitem{mallat}
S.~Mallat.
\newblock {\em A Wavelet Tour of Signal Processing, Third Edition: The Sparse
  Way}.
\newblock Academic Press, 3rd edition, 2008.

\bibitem{subcat}
E.~Ohn-Bar and M.~Trivedi.
\newblock Learning to detect vehicles by clustering appearance patterns.
\newblock {\em TITS}, 2015.

\bibitem{xiaogang1}
W.~Ouyang and X.~Wang.
\newblock A discriminative deep model for pedestrian detection with occlusion
  handling.
\newblock In {\em CVPR}, 2012.

\bibitem{bojan_cvpr12}
B.~Pepik, M.~Stark, P.~Gehler, and B.~Schiele.
\newblock Teaching 3d geometry to deformable part models.
\newblock In {\em CVPR}, 2012.

\bibitem{bojan_cvpr13}
B.~Pepik, M.~Stark, P.~Gehler, and B.~Schiele.
\newblock Occlusion patterns for object class detection.
\newblock In {\em CVPR}, 2013.

\bibitem{yolo}
J.~Redmon, S.~K. Divvala, R.~B. Girshick, and A.~Farhadi.
\newblock You only look once: Unified, real-time object detection.
\newblock In {\em CVPR}, 2016.

\bibitem{faster_rcnn}
S.~Ren, K.~He, R.~Girshick, and J.~Sun.
\newblock {Faster R-CNN: Towards Real-Time Object Detection with Region
  Proposal Networks}.
\newblock In {\em NIPS}, 2015.

\bibitem{imagenet}
O.~Russakovsky, J.~Deng, H.~Su, J.~Krause, S.~Satheesh, S.~Ma, Z.~Huang,
  A.~Karpathy, A.~Khosla, M.~Bernstein, A.~C. Berg, and L.~Fei-Fei.
\newblock {ImageNet Large Scale Visual Recognition Challenge}.
\newblock {\em International Journal of Computer Vision (IJCV)},
  115(3):211--252, 2015.

\bibitem{Schulter_2014_CVPR}
S.~Schulter, C.~Leistner, P.~Wohlhart, P.~M. Roth, and H.~Bischof.
\newblock Accurate object detection with joint classification-regression random
  forests.
\newblock In {\em The IEEE Conference on Computer Vision and Pattern
  Recognition (CVPR)}, June 2014.

\bibitem{overfeat}
P.~Sermanet, D.~Eigen, X.~Zhang, M.~Mathieu, R.~Fergus, and Y.~LeCun.
\newblock Overfeat: Integrated recognition, localization and detection using
  convolutional networks.
\newblock {\em CoRR}, abs/1312.6229, 2013.

\bibitem{ohem}
A.~Shrivastava, A.~Gupta, and R.~Girshick.
\newblock Training region-based object detectors with online hard example
  mining.
\newblock In {\em Conference on Computer Vision and Pattern Recognition
  ({CVPR})}, 2016.

\bibitem{DisAOT_CVPR2013}
X.~Song, T.~Wu, Y.~Jia, and S.-C. Zhu.
\newblock Discriminatively trained and-or tree models for object detection.
\newblock In {\em CVPR}, 2013.

\bibitem{dropout}
N.~Srivastava, G.~Hinton, A.~Krizhevsky, I.~Sutskever, and R.~Salakhutdinov.
\newblock Dropout: A simple way to prevent neural networks from overfitting.
\newblock {\em J. Mach. Learn. Res.}, 15(1), Jan. 2014.

\bibitem{googlenet}
C.~Szegedy, W.~Liu, Y.~Jia, P.~Sermanet, S.~E. Reed, D.~Anguelov, D.~Erhan,
  V.~Vanhoucke, and A.~Rabinovich.
\newblock Going deeper with convolutions.
\newblock In {\em CVPR}, 2015.

\bibitem{szegedy_nips13}
C.~Szegedy, A.~Toshev, and D.~Erhan.
\newblock Deep neural networks for object detection.
\newblock In C.~J.~C. Burges, L.~Bottou, M.~Welling, Z.~Ghahramani, and K.~Q.
  Weinberger, editors, {\em Advances in Neural Information Processing Systems
  26}, pages 2553--2561, 2013.

\bibitem{torralba}
A.~Torralba.
\newblock Contextual priming for object detection.
\newblock {\em IJCV}, 53:2003, 2003.

\bibitem{SS}
J.~R.~R. Uijlings, K.~E.~A. van~de Sande, T.~Gevers, and A.~W.~M. Smeulders.
\newblock Selective search for object recognition.
\newblock {\em International Journal of Computer Vision}, 104(2):154--171,
  2013.

\bibitem{laptev15}
T.~Vu, A.~Osokin, and I.~Laptev.
\newblock Context-aware {CNNs} for person head detection.
\newblock In {\em International Conference on Computer Vision (ICCV)}, 2015.

\bibitem{wanli}
L.~Wan, D.~Eigen, and R.~Fergus.
\newblock End-to-end integration of a convolution network, deformable parts
  model and non-maximum suppression.
\newblock In {\em The IEEE Conference on Computer Vision and Pattern
  Recognition (CVPR)}, June 2015.

\bibitem{3dvp}
Y.~Xiang, W.~Choi, Y.~Lin, and S.~Savarese.
\newblock Data-driven 3d voxel patterns for object category recognition.
\newblock In {\em CVPR}, 2015.

\bibitem{ramananPose}
Y.~Yang and D.~Ramanan.
\newblock Articulated pose estimation with flexible mixtures-of-parts.
\newblock In {\em CVPR}, 2011.

\bibitem{multipath}
S.~Zagoruyko, A.~Lerer, T.-Y. Lin, P.~O. Pinheiro, S.~Gross, S.~Chintala, and
  P.~Dollár.
\newblock A multipath network for object detection.
\newblock In {\em British Machine Vision Conference (BMVC)}, 2016.

\bibitem{yuting}
Y.~Zhang, K.~Sohn, R.~Villegas, G.~Pan, and H.~Lee.
\newblock Improving object detection with deep convolutional networks via
  bayesian optimization and structured prediction.
\newblock In {\em CVPR}, 2015.

\bibitem{zhu_grammar}
S.-C. Zhu and D.~Mumford.
\newblock A stochastic grammar of images.
\newblock {\em Found. Trends. Comput. Graph. Vis.}, 2006.

\bibitem{yukun}
Y.~Zhu, R.~Urtasun, R.~Salakhutdinov, and S.~Fidler.
\newblock segdeepm: Exploiting segmentation and context in deep neural networks
  for object detection.
\newblock In {\em CVPR}, 2015.

\bibitem{edge_boxes}
C.~L. Zitnick and P.~Doll\'ar.
\newblock Edge boxes: Locating object proposals from edges.
\newblock In {\em ECCV}, 2014.

\end{thebibliography}
}

\end{document}